\documentclass[10pt,twocolumn]{article}

\date{}

\usepackage{times}
\usepackage{graphicx}
\usepackage{amsmath}
\usepackage{amssymb}
\usepackage{bbm}
\usepackage{hyperref}

\newcommand{\RCHC}{RCHC (Ours)}

\begin{document}

\title{Reconciling a Centroid-Hypothesis Conflict in Source-Free Domain Adaptation}

\author{Idit Diamant, Roy H. Jennings, Oranit Dror, Hai Victor Habi, Arnon Netzer\\
Sony Semiconductor Israel\\
{\tt\small \{idit.diamant,roy.jennings,oranit.dror,hai.habi,arnon.netzer\}@sony.com}\\
}

\maketitle

\begin{abstract}
Source-free domain adaptation (SFDA) aims to transfer knowledge learned from a source domain to an unlabeled target domain, where the source data is unavailable during adaptation. Existing approaches for SFDA focus on self-training usually including well-established entropy minimization techniques. One of the main challenges in SFDA is to reduce accumulation of errors caused by domain misalignment. A recent strategy successfully managed to reduce error accumulation by pseudo-labeling the target samples based on class-wise prototypes (centroids) generated by their clustering in the representation space. However, this strategy also creates cases for which the cross-entropy of a pseudo-label and the minimum entropy have a conflict in their objectives. We call this conflict the centroid-hypothesis conflict. We propose to reconcile this conflict by aligning the entropy minimization objective with that of the pseudo labels' cross entropy. We demonstrate the effectiveness of aligning the two loss objectives on three domain adaptation datasets. 
In addition, we provide state-of-the-art results using up-to-date architectures also showing the consistency of our method across these architectures. 
The code is available at: \url{https://github.com/ssi-research/SFDA-RCHC}.

\end{abstract}

\begin{figure}[t!]
\centering{\includegraphics[width=1.0\columnwidth]{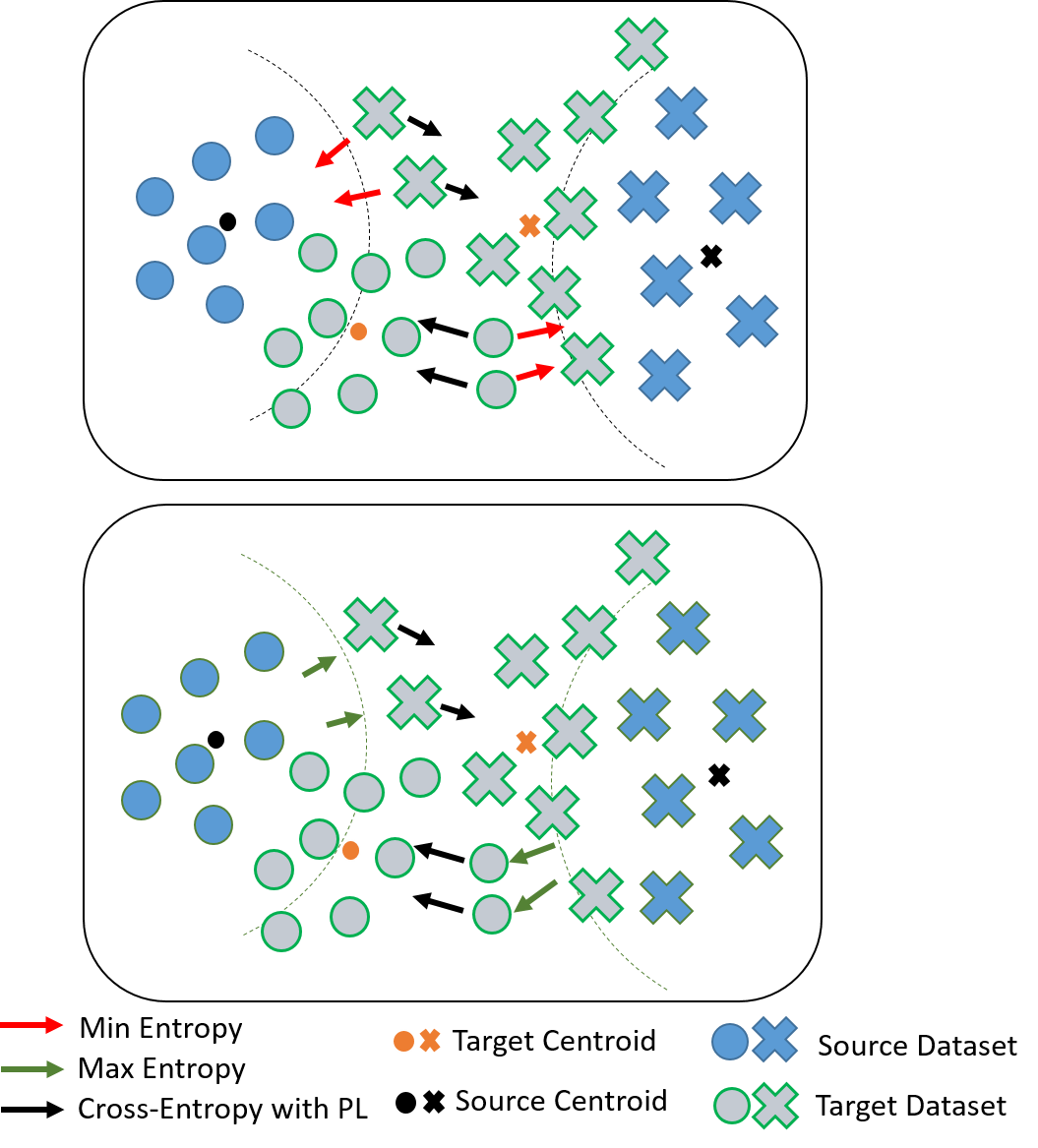}}
\caption{The Centroid-Hypothesis Conflict. 
\textbf{Top}. Illustrates the conflict between two self-training loss terms: hypothesis entropy minimization and pseudo-label cross-entropy. 
\textbf{Bottom}. Illustrates the desired alignment of entropy maximization with the cross-entropy loss.}
\label{Fig:ReconcilingExample}
\end{figure}

\section{Introduction}
\label{sec:intro}
Unsupervised domain adaptation (UDA) adapts a predictive model from a labeled source domain to an unlabeled target domain under possible domain shifts.
Source-free domain adaption (SFDA) is UDA with the added constraint that the source domain data is unavailable during adaptation to the unlabeled target domain. 
As such, SFDA relies entirely on unsupervised learning and self-training techniques.

One of the main challenges in SFDA is to reduce accumulation of errors caused by domain misalignment.
Existing approaches for SFDA focus on self-training using target pseudo-labels and well established entropy minimization techniques \cite{SHOT2020,SHOTplus2021,Inheritne2020,LeeICML2022}.
Specifically, \cite{SHOT2020} managed to successfully reduce error accumulation by assigning pseudo-labels (PLs) to the target data using class clusters in the penultimate layer of the model, where the PLs are based on distance from the centroids of the clusters.

In this work, we focus on two loss terms used in conjunction: the cross-entropy (CE) loss applied to PLs and the entropy minimization loss applied to softmax outputs of the network (the hypotheses).
If the pseudo-label and the softmax output of a target sample disagree, then these two losses have a conflict in their objectives, see Figure \ref{Fig:ReconcilingExample} (top).
The CE loss increases the probability of the class implied by the PL and the entropy minimization loss increases the probability of the class implied by the hypothesis.
We call this conflict the \textit{\textbf{centroid-hypothesis conflict (CHC)}}. 

We suggest to reconcile CHC for samples where we have high certainty in the loss term implied by the PL.
In order to identify these samples, we introduce a metric based on target class centroids and reconcile CHC by replacing entropy minimization with maximization, which is better aligned with the CE loss, see Figure \ref{Fig:ReconcilingExample} (bottom).

Our main \textbf{contributions} are as follows:

\begin{itemize}
    \item We introduce a metric measuring the uncertainty of pseudo label assignments based on target class centroids.
    \item We apply the above metric to introduce a strategy to reconcile the centroid-hypothesis conflict (RCHC) between self-supervised loss terms based on pseudo labels and entropy minimization, aligning their objectives.
    \item We demonstrate the effectiveness of aligning the loss objectives on VisDA, DomainNet and OfficeHome.
    \item We apply RCHC to both ResNet and ConvNeXt backbones showing its effectiveness across up-to-date architectures including backbones pre-trained on ImageNet-22K.
    \item To the best of our knowledge, we are the first to present SFDA results using ConvNeXt, obtaining state-of-the-art results on VisDA, DomainNet and OfficeHome datasets.
\end{itemize}

\section{Related Work}

\subsection{Unsupervised Domain Adaptation}
Unsupervised domain adaptation (UDA) 
is receiving much attention in recent years \cite{DANN2016, Morerio2018, BN_DA_CVPR2019, SENTRY2020, Li2020RethinkingDM,ADDA2017,HashingDA2017,UniversalDA2019,saito2017maximum, DirtUDA2018,Mitsuzumi_2021_CVPR}. One common approach to tackle UDA is distributional matching (DM), which can be applied either by directly minimizing the domain discrepancy statistics \cite{sun2016deep, Morerio2018} or via domain-adversarial learning \cite{DANN2016}. 
However, \cite{Li2020RethinkingDM} showed that DM-based approaches have limited generalization ability under what they coin "realistic shifts" in which the source label distribution is balanced and the target label distribution is long tailed.
The authors showed that self-training methods are more robust for UDA. 
In \cite{SENTRY2020}, it has been shown that self-training methods achieve promising results for UDA. 
The authors identified reliable and unreliable target instances based on their predictive consistency under a committee of random transformations and applied selective entropy optimization.

\subsection{Source-Free Domain Adaptation}

Recently, due to awareness of privacy and data protection, there is an increasing interest in the source-free scenario of domain adaptation \cite{SHOT2020,yang2022attracting,A2Net2021, Li_2020_CVPR,sanqing2022BMD,Qiu2021CPGA}.
SFDA approaches are based on self-training and unsupervised techniques due to the absence of labeled data.
\cite{yang2022attracting} treated SFDA as an unsupervised clustering problem and proposed to utilize prediction consistency. 
The intuition behind this approach is that local neighbors in the feature space should have more similar predictions than other neighbors.
\cite{A2Net2021} suggested an adaptive adversarial network that is combined with a contrastive category-wise matching module and self-supervised rotation loss.
In \cite{SFDADE2022}, 
surrogate features were sampled from an estimated source distribution and were utilized to align the two domains by minimizing a contrastive adaptation loss function.
\cite{SHOT2020,SHOTplus2021} utilized mutual information maximization and cross-entropy with pseudo-labels to adapt a source model to a target domain.
The approach used in this work gave rise to the centroid-hypothesis conflict, which our work resolves.

\section{Preliminaries}
\label{section:preliminaries}
In this work, we study the problem of source-free unsupervised domain adaptation (SFDA) for the visual classification task.
In this scenario, a model is adapted to an unlabeled target domain using only the pre-trained source model with no access to source data.
Throughout this work, we deal only with a single source domain and the closed set scenario, in which the source and target domains share the same set of classes.

\subsection{SFDA Problem Setup}

We study SFDA in the context of a $K$-way image classification task.
The problem is usually formulated as a two-stage training procedure. 
In the first stage, we learn a predictive function $h_s: \mathcal{X}_s \rightarrow \mathcal{Y}$ using $n_s$ labeled samples $\{x^i_s, y^i_s\}_{i=1}^{n_s}$ from a source domain $\mathcal{D}_s$ where $x^i_s \in \mathcal{X}_s$ and $y^i_s \in \mathcal{Y}$ are the samples and their associated labels respectively.
Usually, $h_s(x_s)=\mathrm{argmax}(g_s(f_s(x_s)))$ where $f_s$ is a feature extractor initialized using a pre-trained (on ImageNet) model and $g_s$ a task specific classification head.

In the second stage, we are given $n_t$ unlabeled samples $\{x_t^i\}_{i=1}^{n_t}$ from a target domain $\mathcal{D}_t$ where $x_t\in \mathcal{X}_t$ and train a target function $h_t: \mathcal{X}_t \rightarrow \mathcal{Y}$ to predict the labels $\{y_t^i\}_{i=1}^{n_t}$ of the target samples using only the source function $h_s$ and $\mathcal{X}_t$ (with no access to source data).
We denote $h_t(x_t)=\mathrm{argmax}(g_t(f_t(x_t)))$, similarly to $h_s$.

\begin{figure*}[t!]
\centering{\includegraphics[width=0.9\textwidth]{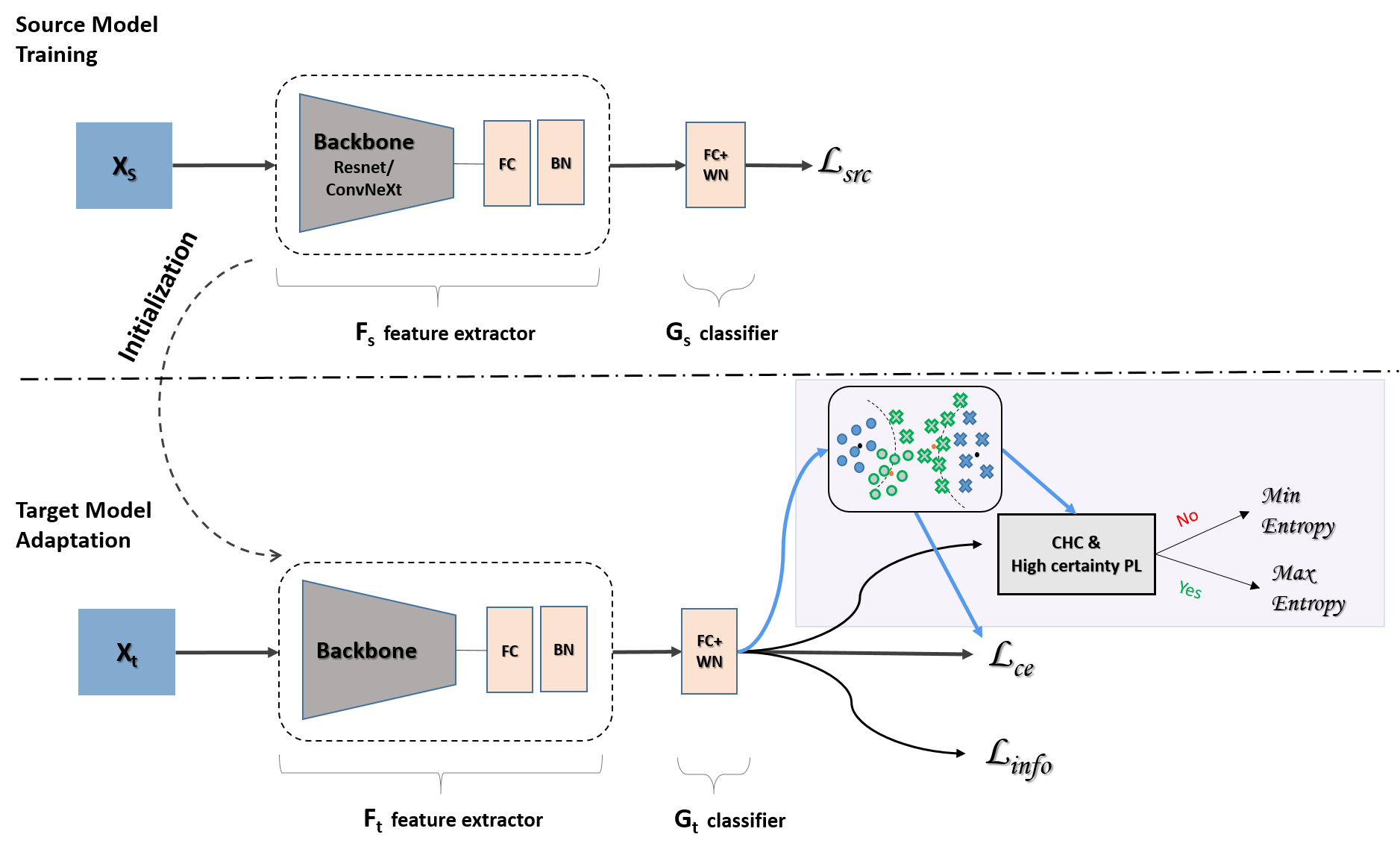}}
\caption{Overview of the proposed method reconciling the centroid-hypothsis conflict.
Target class-centroids are used both for pseudo-label generation and for aligning the entropy minimization objective with that of the pseudo labels' cross entropy.} 
\label{Fig:Target_Model}
\end{figure*}

\subsection{Source Model Training}

The source classification model $h_s$
is learned in a standard supervised fashion based on cross-entropy with label smoothing \cite{SHOT2020}:

\begin{equation}
\begin{aligned}
& \mathcal{L}_{src}\, =\, \mathbb{E}_{(x_s, y_s) \in \mathcal{D}_s } \sum^K_{k=1}\, -\hat{q}_{k}\; log(\, \sigma_k(g_s(f_s(x_s)))),
\end{aligned}
\label{eq:gan}
\end{equation}

\noindent
where $\hat{q}_k=\frac{\alpha}{K}+(1-\alpha)q_k$ is the $k$-th entry of the smoothed label corresponding to $y_s$, $\alpha$ is the smoothing parameter, $q_{k}$ is the $k$-th entry of the one-hot encoding of $y_s$ and
$\sigma_k$ is the \textit{k}-th element of the softmax function's output.

\subsection{Target Model Adaptation}
\label{par:Target_Adaptation}
We follow the adaptation process of SHOT in \cite{SHOTplus2021}, which includes a rotation loss (see Equation \ref{eq:rotation_loss}).
During target adaptation, the classifier $g_t$ is initialized with the source classifier's weights and is frozen during fine-tuning of the feature encoder. 
The following paragraphs describe the losses that are imposed during target adaptation.

The model is learned by maximizing the \textit{{\textbf{mutual information}}} \cite{Li2020RethinkingDM, SHOT2020, Hu_IM_2017} between the input data and the classifier's predictions. 
The target's mutual information can be expressed as: $I(\mathcal{X}_t;\mathcal{P}_t)=\mathrm{H}(\mathcal{P}_t)-\mathrm{H}(\mathcal{P}_t|\mathcal{X}_t)$ where $\mathcal{P}_t = h_t(\mathcal{X}_t)$ is the set of target predictions (hypotheses). 
Mutual information maximization is obtained by maximizing the \textit{\textbf{info-entropy}} $\mathrm{H}(\mathcal{P}_t)$ and minimizing the \textit{\textbf{conditional entropy}} $\mathrm{H}(\mathcal{P}_t|\mathcal{X}_t)$. 
Minimizing the conditional entropy pushes the target data away from the decision boundaries while maximizing the info-entropy insures the diversity of the output (preventing the degenerate solution of a single class prediction). 
The mutual \textbf{\textit{information maximization}} loss is expressed as follows:

\begin{equation}
\mathcal{L}_{im}(g_t, f_t;\mathcal{X}_t) = \mathcal{L}_{info}+\mathcal{L}_{H},
\label{eq:IM}
\end{equation}
where the first term $\mathcal{L}_{info}$ is the info-entropy loss $\sum^K_{k=1} \, \hat{p}_k \log \hat{p}_k$ with $\hat{p}=\mathbb{E}_{x_t \in \mathcal{X}_t}\sigma(g_t(f_t(x_t)))$ and the second loss term $\mathcal{L}_{H}$ is the conditional entropy $-\mathbb{E}_{x_t \in \mathcal{X}_t } \sum^K_{k=1}\, \sigma_k(g_t(f_t(x_t)))\log(\sigma_k(g_t(f_t(x_t))))$.

Entropy minimization $\mathcal{L}_{H}$ encourages target feature representations to fit the source classifier well.
This might result in error accumulation during adaptation when matching target representations with wrong (source based) hypotheses.
To alleviate this problem, cross-entropy (CE) using pseudo-labels based on the target distribution is applied.
The pseudo-labels are generated as follows. 
First, the centroid of each class in the target domain is obtained with weighted clustering 
$C_k = \frac{\sum_{x_t \in \mathcal{X}_t}  \sigma_k(g_t(f_t(x_t)))f_t(x_t)} {\sum_{x_t \in \mathcal{X}_t}  \sigma_k(g_t(f_t(x_t)))}$.
Then, for each target instance, a pseudo-label is generated based on its nearest centroid using cosine distance.
Finally, the process is repeated one more time: the target class centroids $C_k$ are refined based on the pseudo-labels and the final pseudo-labels 
are computed using the refined centroids. 
We denote the final pseudo-labeled target domain $\{x_t, \hat{y}_t\}_{i=1}^{n_t}$ by $\hat{\mathcal{D}}_t$.
The cross-entropy loss with the pseudo-labels is defined as follows:

\begin{equation}
\mathcal{L}_{ce} = 
- \mathbb{E}_{(x_t,\hat{y}_t)\in \hat{\mathcal{D}}_t}
\sum^{K}_{k=1} \mathbbm{1}_{[\hat{y}_t = k]}  \log \sigma_{k}(g_t(f_t(x_t))).
\label{eq:CE-PL}
\end{equation}

\textit{\textbf{Rotation classification}} is an unsupervised task defined to predict an arbitrary relative 2D rotation $r \in \mathcal{R}$ for  $\mathcal{R} = \{0^\circ, 90^\circ, 180^\circ, 270^\circ\}$ of an image $x$, given both $x$ and it's rotated version $x^{r}$.
For this purpose, a rotation classifier head $g_c$ is added to the network with four outputs corresponding to the four rotation degrees (using the same feature encoder $f_t$ as the main classification head $g_t$). 
The probability of the \textit{k}-th relative rotation degree predicted by $g_c$ is given by $\sigma_{k}\, (g_c([f_t(x_t),f_t(x^{r}_t)])) $.
The rotation loss is defined as follows:

\begin{equation}
\begin{aligned}
& \mathcal{L}_{rot}\, (f_t, g_c; \mathcal{X}_t, \mathcal{R})= \\ 
& -\mathbb{E}_{(x_t,r)\in \mathcal{X}_t \times \mathcal{R}} \sum^{4}_{k=1}{\mathbbm{1}_{[r=k]}} \log \sigma_{k} ( g_c(\, [f_t(x_t),f_t(x^{r}_t)])).
\end{aligned}
\label{eq:rotation_loss}
\end{equation}

\section{Method}

In this section we present the centroid-hypothesis conflict, a conflict between objectives of two self-supervised loss functions used for SFDA.
We suggest a strategy to combine the two loss functions in a way that reconciles the conflict.
We base our solution on the training procedure described in Section \ref{section:preliminaries} and update $\mathcal{L}_{\mathrm{H}}$ in the process. 
For an overview of our proposed strategy, see Figure \ref{Fig:Target_Model}.

\subsection{The Centroid-Hypothesis Conflict}
In SFDA, adaptation of a model to a target domain encourages the model to make confident predictions on unlabeled target data.
One of the main challenges in SFDA is to reduce accumulation of errors caused by domain misalignment.
Recent SFDA methods \cite{SHOT2020, SHOTplus2021, LeeICML2022}, use both well established entropy minimization techniques and cross-entropy (CE) with target pseudo-labels generated using information in representation clusters of the penultimate layer of the model. That is, the pseudo-labels (PLs) of target samples are determined by their distance from the centroids of class clusters.

Cross-entropy loss using target PLs and entropy minimization loss might have contradictive objectives.
That is, CE loss increases the probability of the class implied by the PL and the entropy minimization loss increases the probability of the class implied by the hypothesis.
We call this conflict the \textbf{centroid-hypothesis conflict} (CHC).

Figure \ref{Fig:ReconcilingExample} illustrates CHC in a two-class scenario. 
The two classes are denoted by $\circ$ and $\times$. 
The figure illustrates four examples of misaligned target instances, a pair from each category. These instances are pulled in different directions by the cross-entropy and entropy minimization loss functions.
The entropy minimization pulling towards the centroids of incorrect categories, while the cross-entropy is pulling towards the respective PLs.

\subsection{Reconciling the Centroid-Hypothesis Conflict}
\label{sec:RCHC}
We suggest to reconcile the conflict by replacing entropy minimization with maximization which is better aligned with the CE loss, see Figure \ref{Fig:ReconcilingExample} (bottom).
However, to overcome the existence of noise in the pseudo-labels, we reconcile samples with high certainty in the pseudo-label assignment.
For samples with low certainty, we apply both entropy minimization and cross-entropy with pseudo-labels which increases regularization to the optimization process. 

We introduce a metric, ${r}(z) \in [0,1]$, that measures the uncertainty of the PL assignment of a sample based on its distance from target class centroids as follows:

\begin{equation}
{r}(z) = \frac{\min\limits_{k\in K} \;D(z,c_k)}{{{\min\limits_{k\in K}}^{2\rm{nd}}} \; D(z,c_k)},
\label{Eq:Ratio}
\end{equation}
where $z$ is the target sample embedding, $c_k$ is a target centroid, $D(\cdot,\cdot)$ is the cosine distance and $\min^{2nd}$ is the second minimal element.

\begin{figure}[t!]
\centering{\includegraphics[width=0.9\columnwidth]{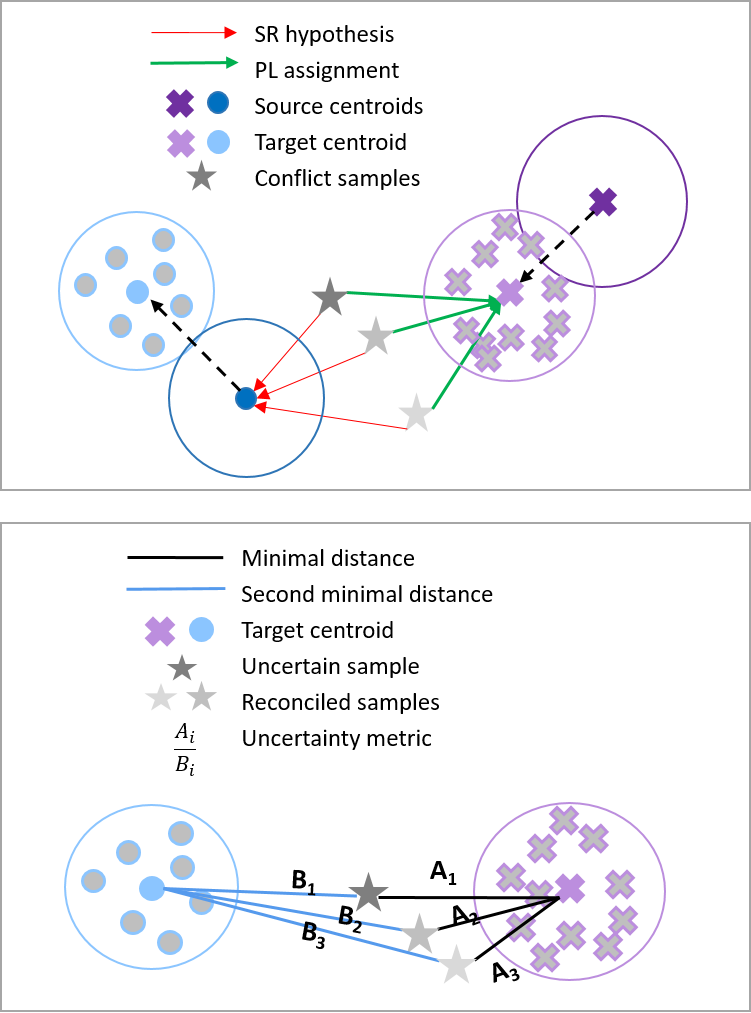}}
\caption{Top. Illustration of conflict samples based on their softmax-response (SR) hypotheses and pseudo-label (PL) assignments; Bottom. Geometric visualisation of the uncertainty metric definition.}
\label{Fig:RatioExplanation}
\end{figure}

Figure \ref{Fig:RatioExplanation} illustrates the rational behind our uncertainty metric. 
The three samples (marked with $\star$) in the illustrations demonstrate that a sample that is close to one centroid and far away from all other centroids is more likely to have a correct pseudo-label.
In contrast, as a sample gets closer to a boundary between clusters, the uncertainty of its pseudo-label assignment increases.

We use $r$ in order to define a conditional entropy loss $\mathcal{L}_{\mathrm{ent}}$ and replace $\mathcal{L}_{H}$ in Equation \ref{eq:IM} with $\mathcal{L}_{\mathrm{ent}}$ obtaining our adaptation strategy Reconciling the Centroid-Hypothesis Conflict (RCHC).

In order to define $\mathcal{L}_{\mathrm{ent}}$, we first need to define the subset $C_{chc}$ of target samples that hold CHC and in which we have high certainty in the assigned PLs:
\begin{equation}
\begin{aligned}
 C_{chc}=\{x_t :\;&\hat{y}_t\, \neq  \mathrm{argmax} \; \sigma(g_t(f_t(x_t)))\;\text{and}\\
 &r(f_t(x_t)) < r^{th}\}
 \end{aligned}
\end{equation}

\noindent
where $x_t$ is a target sample and $\hat{y}_t$ its associated pseudo-label,
$\sigma$ is the softmax function's output,  and $r^{th}$ is a threshold hyper-parameter.
We define the conditional entropy loss $\mathcal{L}_{\mathrm{ent}}$ as follows:

\begin{equation}
\mathcal{L}_{\mathrm{ent}} = \mathbb{E}_{x_t \in \mathcal{X}_t}\delta(x_t)\mathrm{H}(y|x_t),
\end{equation}
where
\begin{equation*}
\begin{aligned}
& \mathrm{H}(y|x_t) = -\sum^K_{k=1}\sigma_k(g_t(f_t(x_t)))\, log\, (\sigma_k(g_t(f_t(x_t)))),\\
&\mathbbm{\delta}(x_t)\, =
 \begin{cases}
    -1& x_t \in C_{chc} \\
    \;\;1   & \text{otherwise}
\end{cases},\\
\end{aligned}
\label{eq:selective_entropy}
\end{equation*}

\noindent
and $\sigma_k$ is the \textit{k}-th element of the softmax function's output. 
Our overall loss is:

\begin{equation}
 \mathcal{L}\, = \mathcal{L}_{\mathrm{ent}} + \mathcal{L}_{\mathrm{info}}  + \alpha\, \mathcal{L}_{ce} + \beta \, \mathcal{L}_{\mathrm{rot}},
\label{eq:selective_entropy_full_loss}
\end{equation}

\noindent
where $\alpha$ and $\beta$ are balancing hyper-parameters.

\section{Experiments}

In this section, we describe our experimental setup and present results for three different domain adaptation benchmarks: OfficeHome, ViSDA and DomainNet.
We provide results based both on ResNet and ConvNeXt \cite{liu2022convnet} architectures.

Recently, \cite{kim2022unified} implied that adaptation strategies applied to ConvNeXt pre-trained on ImangeNet-22K might result with negative transfer on some of the domain shifts. 
Additionally, they raised a concern that there is a need to compare adaptation methods not only on ResNet-based backbones but also for new architectures, since the methods ranking might be different.
We therefore extend our results to ConvNext architectures pre-trained on ImageNet-22K to examine our method's effectiveness and consistency across architectures over all of the domain shifts.

To the best of our knowledge, we are the first to present SFDA results with ConvNeXt for ViSDA, OfficeHome and DomainNet using variants of ConvNeXt, obtaining state-of-the-art results with model sizes comparable to those in previous SOTA results.

\subsection{Datasets}

\paragraph{OfficeHome} \cite{HashingDA2017} is an image classification benchmark containing 65 categories of objects. 
It consists of four distinct domains: Art (A), Clipart (C), Product (P) and RealWorld images (R).
\paragraph{VisDA} \cite{peng2017visda} is a large-scale benchmark for synthetic-to-real adaptation. 
It contains over 200K images of 12 classes.
\paragraph{DomainNet} \cite{peng2019moment} is a large UDA benchmark for image classification. Since its full version suffers from labeling noise, we use the DomainNet mini subset proposed in \cite{TanECCV2020}. This subset contains 40 classes from 4 domains: Real (R), Clipart (C), Painting (P) and Sketch (S).  
\paragraph{Metrics.} To be consistent with prior art, for VisDA and DomainNet we use the per-class accuracy metric, while on OfficeHome we report the standard (overall) accuracy.

\begin{table*}[th!]
\centering
\resizebox{1.0\textwidth}{!}{%
    \begin{tabular}{@{}llllllllllllllll@{}} \hline
     Method (Synthesis→Real) & Backbone & Params & Plane & Bcycle &  Bus & Car &  Horse & Knife & Mcycl & Person & Plant & Sktbrd & Train & Truck & AVG\\\hline

    Source only  & Resnet-101& 42M& 60.9 & 21.6 & 50.9 & 67.6 & 65.8 & 6.3 & 82.2 & 23.2 & 57.3 & 30.6 & 84.6 & 8.0 & 46.6 \\ 
    A$^2$Net \cite{A2Net2021} & Resnet-101& 42M& 94.0& 87.8 &85.6& 66.8& 93.7& 95.1& 85.8& 81.2& 91.6 &88.2& 86.5 &56.0& 84.3\\
    G-SFDA \cite{GSFDA2021} & Resnet-101& 42M& 96.1 &88.3 &85.5& 74.1& 97.1 &95.4& 89.5& 79.4& 95.4 &92.9& 89.1& 42.6& 85.4\\
    SFDA-DE \cite{SFDADE2022}& Resnet-101& 42M& 95.3 &91.2& 77.5& 72.1 &95.7& 97.8 &85.5& 86.1& 95.5 &93.0& 86.3& 61.6& 86.5\\
    CoWA-JMDS \cite{LeeICML2022} & Resnet-101& 42M & 96.2 &89.7& 83.9& 73.8& 96.4& 97.4& 89.3& 86.8 &94.6 &92.1 &88.7 &53.8 & 86.9 \\
    AaD \cite{yang2022attracting}& Resnet-101& 42M & 97.4 &90.5 &80.8 &76.2 &97.3 &96.1 &89.8& 82.9 &95.5& 93.0 &92.0 &64.7& 88.0\\
     \hline
    SHOT \cite{SHOTplus2021}& Resnet-101& 42M & 95.8& 88.2& 87.2& 73.7& 95.2& 96.4& 87.9&84.5& 92.5& 89.3& 85.7& 49.1& 85.5\\  
    \textbf{\RCHC}& Resnet-101& 42M  & 96.0 & 90.1 & 85.0 & 73.1 & 95.9 & 97.3 & 87.6 & 84.7 & 93.8 & 91.3 & 86.7 & 51.2 &86.1 \\
     \hline
    SHOT++ \cite{SHOTplus2021}& Resnet-101& 42M  & 97.7  &88.4  &90.2  &86.3  &97.9 & 98.6  &92.9 & 84.1 &97.1  &92.2 & 93.6  &28.8 & 87.3\\
    \textbf{\RCHC++} & Resnet-101& 42M & 97.6 & 88.9 & 88.4 & 84.0 & 97.6 & 97.4 & 92.2 & 86.2 & 97.4 & 92.8 & 92.6 & 41.2 & 87.8 \\
     \hline
     \hline
      Source only & ConvNeXt-S& 49M  & 98.4 & 67.6 & 78.8 & 73.4 & 84.1 & 49.1 & 94.0 & 10.0 & 65.8 & 89.7 & 94.8 & 16.9 & 68.6\\
    SHOT & ConvNeXt-S& 49M  & 97.9 & 93.8 & 86.2 & 75.9 & 97.4 & 98.6 & 92.0 & 76.2 & 94.4 & 96.0 & 92.9 & 67.3 & 89.0 \\   
    \textbf{\RCHC}  & ConvNeXt-S& 49M & 98.4 & 94.3 & 85.2 & 77.4 & 98.0 & 98.8 & 92.1 & 79.5 & 94.7 & 96.0 & 93.0 & 64.8 & \textbf{89.3} \\
     \hline
     \hline
      Source only & ConvNeXt-T& 27M  & 97.0 & 61.6 & 71.7 & 77.9 & 87.7 & 67.3 & 91.9 & 13.9 & 64.0 & 82.3 & 93.7 & 20.4 & 69.1\\
    SHOT & ConvNeXt-T& 27M & 97.4 & 91.2 & 82.6 & 70.3 & 96.6 & 98.1 & 89.7 & 79.6 & 94.0 & 94.1 & 90.9 & 61.7 & 87.2 \\
    \textbf{\RCHC}&ConvNeXt-T& 27M & 97.5 & 91.6 & 82.7 & 71.3 & 96.7 & 98.1 & 89.6 & 80.6 & 94.2 & 94.4 & 90.9 & 60.0 & 87.3 \\

 \hline
    \end{tabular}
}
\caption {Performance comparison for VisDA with both Resnet and ConvNeXt-22K backbones.}
\label{tab:visda-performance}
\end{table*}


\begin{table*}[ht!]
 \centering
\resizebox{1.0\textwidth}{!}{%
    \begin{tabular}{@{}llllllllllllllll@{}} \hline
     Method & Backbone & Params &  $A\xrightarrow[]{}C$ & $A\xrightarrow[]{}P$ & $A\xrightarrow[]{}R$ & $C\xrightarrow[]{}A$ & $C\xrightarrow[]{}P$ & $C\xrightarrow[]{}R$ & $P\xrightarrow[]{}A$ & $P\xrightarrow[]{}C$ & $P\xrightarrow[]{}R$ & $R\xrightarrow[]{}A$ & $R\xrightarrow[]{}C$ & $R\xrightarrow[]{}P$ & AVG\\\hline 
    Source only  & Resnet-50& 23M   & 44.6 & 67.3 &74.8 & 52.7 &62.7 & 64.8 & 53.0 &40.6 & 73.2 & 65.3 & 45.4 & 78.0 & 60.2 \\ 

     A$^2$Net \cite{A2Net2021}& Resnet-50& 23M  & 58.4 &79.0 &82.4& 67.5& 79.3 &78.9 &68.0 &56.2 &82.9 &74.1& 60.5& 85.0 &72.8\\
    
     SFDA-DE \cite{SFDADE2022} & Resnet-50& 23M & 59.7& 79.5& 82.4 &69.7& 78.6& 79.2& 66.1 &57.2 &82.6& 73.9 &60.8& 85.5 &72.9\\
     CoWA-JMDS \cite{LeeICML2022} & Resnet-50& 23M  & 56.9 & 78.4& 81.0& 69.1& 80.0& 79.9& 67.7& 57.2& 82.4 &72.8 &60.5 &84.5& 72.5 \\
     AaD \cite{yang2022attracting} & Resnet-50& 23M &59.3 &79.3& 82.1 &68.9& 79.8& 79.5& 67.2 &57.4& 83.1& 72.1 &58.5& 85.4 &72.7\\

      \hline
     SHOT* \cite{SHOTplus2021}& Resnet-50& 23M  & 57.2 & 78.8 & 81.3 & 67.9 & 78.5 & 77.8 & 67.4 & 56.1 & 81.8 & 73.3 & 59.9 & 84.6 & 72.0  \\ 
    \textbf{\RCHC}  & Resnet-50& 23M & 57.1 & 78.8 & 81.4 & 68.0 & 78.8 & 78.0 & 67.5 & 55.7 & 81.8 & 73.1 & 59.4 & 84.5 & 72.0 \\
     
       \hline
     SHOT++* \cite{SHOTplus2021} & Resnet-50& 23M & 57.9 & 79.5 & 82.4 & 68.3 & 79.9 & 79.2 & 68.3 & 56.9 & 82.9 & 73.7 & 60.6 & 85.5 & 72.9  \\
    \textbf{\RCHC} ++&  Resnet-50& 23M& 57.8 & 79.7 & 82.4 & 68.6 & 80.3 & 79.3 & 68.1 & 56.9 & 82.6 & 73.6 & 60.0 & 85.3 & 72.9\\
    \hline
    \hline
    Source only  & ConvNeXt-T& 27M  & 63.58 & 81.62 & 86.49 & 75.88 & 80.98 & 83.28 & 73.11 & 59.65 & 87.14 & 79.19 & 62.39 & 87.84 & 76.76\\
     SHOT & ConvNeXt-T& 27M  & 73.20 & 88.83 & 90.31 & 84.85 & 88.80 & 89.73 & 82.06 & 71.45 & 90.21 & 84.08 & 73.58 & 91.46 & 84.05  \\ 
   \textbf{\RCHC}  & ConvNeXt-T& 27M & 72.95 & 88.87 & 90.12 & 84.91 & 88.56 & 89.71 & 82.57 & 71.53 & 90.47 & 84.34 & 73.30 & 91.57 & \textbf{84.08} \\
    \hline
       \end{tabular}
    }
 \caption {Performance comparison for OfficeHome with both  Resnet and ConvNeXt-22K backbones. We denote by * results reproduced using our implementation.}
  \label{tab:OH-performance}
\end{table*}

\begin{table*}[th!]
 \centering
\resizebox{1.0\textwidth}{!}{%
    \begin{tabular}{@{}llllllllllllllll@{}} \hline
     Method & Backbone & Params &   $R\xrightarrow[]{}C$ & $R\xrightarrow[]{}P$ &
     $R\xrightarrow[]{}S$ & $C\xrightarrow[]{}R$ & $C\xrightarrow[]{}P$ & $C\xrightarrow[]{}S$ & $P\xrightarrow[]{}R$ & $P\xrightarrow[]{}C$ & $P\xrightarrow[]{}S$ & $S\xrightarrow[]{}R$ & $S\xrightarrow[]{}C$ &  
     $S\xrightarrow[]{}P$ & AVG\\\hline 
    Source only   & Resnet-50& 23M & 66.5 & 70.4 & 59.5 & 74.6 & 57.1 & 56.7 & 84.5 & 60.9 & 64.4 & 74.4 & 59.7 & 57.8 & 65.6 \\
    SHOT & Resnet-50& 23M & 77.3 & 76.1 & 73.7 & 88.3 & 73.1 & 77.3 & 90.0 & 77.1 & 76.9 & 88.8 & 79.1 & 70.4 & 79.0 \\
    \textbf{\RCHC} & Resnet-50& 23M & 78.2 & 77.2 & 73.7 & 88.4 & 74.0 & 77.1 & 90.2 & 77.2 & 77.2 & 88.8 & 79.4 & 71.3 & 79.4 \\
     \hline
    SHOT++   & Resnet-50& 23M  & 78.2 & 77.5 & 73.9 & 89.1 & 75.5 & 76.4 & 90.6 & 78.2 & 77.3 & 89.3 & 80.0 & 70.9 & 79.7 \\
    \textbf{\RCHC}++ & Resnet-50& 23M & 79.6 & 78.1 & 74.2 & 88.9 & 75.7 & 77.6 & 90.7 & 78.8 & 76.9 & 89.1 & 80.3 & 71.8 & 80.1 \\
   \hline
   \hline
    Source only & ConvNeXt-T& 27M  & 75.87 & 77.69 & 71.13 & 91.80 & 78.29 & 79.36 & 92.30 & 77.87 & 81.52 & 90.81 & 79.77 & 78.88 & 81.27\\
    SHOT& ConvNeXt-T& 27M& 87.31 & 81.90 & 83.23 & 94.14 & 82.26 & 87.01 & 93.44 & 87.39 & 86.92 & 93.06 & 87.97 & 82.87 & 87.29\\
      \textbf{\RCHC}  & ConvNeXt-T& 27M& 88.30 & 82.53 & 83.22 & 94.17 & 82.55 & 86.14 & 93.44 & 87.15 & 86.72 & 93.21 & 87.99 & 83.72 & \textbf{87.43}\\
    \hline
       \end{tabular}
    }
 \caption {Performance comparison for DomainNet (mini) with both Resnet and ConvNeXt-22K backbones.}
  \label{tab:DomainNet-performance}

\end{table*}

\subsection{Implementation Details}
In all of our experiments we use PyTorch and train on an NVIDIA V100 GPU.
We use pre-trained ResNet \cite{he2016deep} and ConvNeXt architectures \cite{liu2022convnet} as backbones. 
Specifically, Resnet50 is used for OfficeHome and DomainNet, while Resnet-101 is used for VisDA.
For each dataset we also provide results using one of ConvNeXt's variants pre-trained on ImageNet-22K, ConvNeXt-T (tiny) or ConvNeXt-S (small), with number of parameters comparable to those in ResNet variant used.

Following \cite{SHOT2020,SHOTplus2021}, we replace the original fully connected layer with a bottleneck layer followed by batch normalization and a task-specific fully connected classifier with weight normalization (see Figure \ref{Fig:Target_Model}). 
We follow the source model training procedure applied in \cite{SHOT2020,SHOTplus2021}. 
Specifically, we train the whole network through back-propagation, the newly added layers trained with a learning rate 10 times that of the pre-trained layers. 
We use mini-batch SGD with momentum 0.9 and weight decay 1e-3.
The learning rate is set to 1e-2 for new layers in all of the experiments, except for VisDA, in which we use a learning rate of 1e-3.

For target adaptation, we update the pseudo-labels at the beginning of each epoch and set the number of epochs to 15. 
All of our results report the mean accuracy over three runs with different random seeds (2019, 2020 and 2021) as was done is \cite{SHOT2020}. 
We set the uncertainty metric threshold hyper-parameter to 0.65 in all our experiments. 
This value is derived from the uncertainty metric statistics for non-conflict points in the beginning of target adaptation which is similar in all of the datasets (see Section \ref{sec:ablation}).

\subsection{Main Results}
\label{sec:MainResults}
In this section, we report results of RCHC on all of the datasets. 
In all tables, "source only" stands for the accuracy of the source model (trained with the source dataset alone) on the target dataset. 
In all results, "++" denotes applying a second fine-tuning stage with MixMatch \cite{MixMatch2019} similarly to \cite{SHOTplus2021}.
SHOT in our tables represents results obtained with the SHOT implementation from \cite{SHOTplus2021} which includes a rotation loss (see Equation \ref{eq:rotation_loss}).

\begin{figure}[t!]
\centering{\includegraphics[width=0.9\columnwidth]{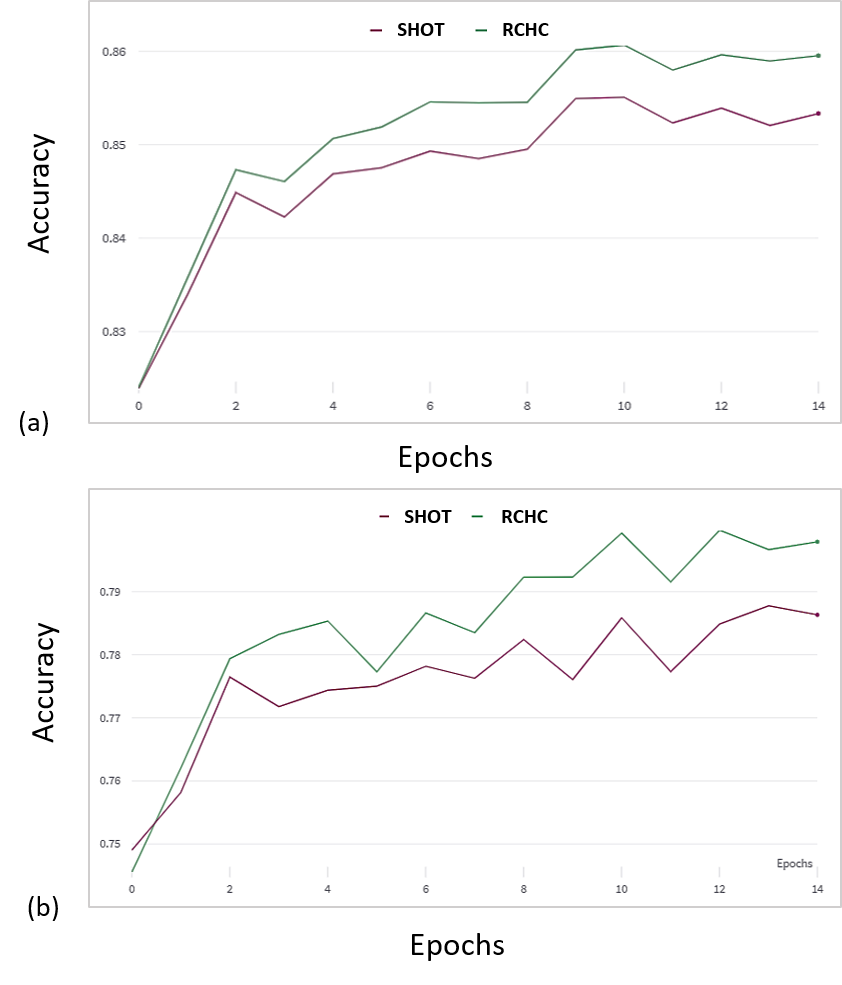}}
\caption{Performance comparison between SHOT and RCHC during adaptation with ResNet backbones; (a) ViSDA - synthetic to real (b) DomainNet - realworld to clipart.} 
\label{Fig:Training_plot}
\end{figure}

Tables \ref{tab:visda-performance} -- \ref{tab:DomainNet-performance} show that RCHC results with consistent improvement when applied to SHOT and SHOT++ with ResNet backbones for all of the datasets.
We also add performance comparison between RCHC and other recent SFDA methods for the VisDA and OfficeHome datasets.
Note that, to the best of our knowledge, we are the first to present SFDA results for the DomainNet mini dataset where we also established the SHOT baseline for comparison.
Figure \ref{Fig:Training_plot} presents examples of performance comparisons between RCHC and SHOT during adaptation with ResNet for ViSDA and DomainNet's realworld to clipart. It shows that the accuracy obtained by RCHC is consistently higher than SHOT's accuracy throughout the training.

The tables also provide results showing consistent improvement over SHOT using ConvNext architectures pre-trained with ImageNet-22K.
Our results show no negative transfer, opposed to the results presented in \cite{kim2022unified}.
That is, RCHC's final adaptation performance is better than source only results for all domain shifts.
The results obtained by ConvNeXt with RCHC are current state-of-the-art on all three datasets using ConvNeXt variants with comparable parameter counts to the ResNet variants used in the literature. 
For ViSDA we obtain accuracy of 89.3\% with ConvNeXt-S, for OfficeHome and DomainNet we obtained accuracy of 84.08\% and 87.43\% respectively with ConvNeXt-T. 

It is interesting to note that RCHC obtains an accuracy of 87.3\% with ConvNeXt-T on ViSDA which improves over resnet-101 by 1.2\% in spite its significantly smaller size.
OfficeHome is unique compared to our other results, in that although RCHC does not impair results it also does not improve them.
We speculate that this is due to higher rate of noisy PLs resulting from ineffective clustering applied to smaller datasets.

\subsection{Ablation Study}
\label{sec:ablation}

In this section, we analyze our choice of the uncertainty metric threshold and provide geometric visualizations validating the behavior of samples holding the centroid-hypothesis conflict.

\paragraph{Threshold selection.}
The uncertainty metric values are in the range $[0,1]$. Setting the threshold $r^{th}=0$ is identical to applying SHOT without RCHC, always using both losses. 
Setting the threshold $r^{th}=1$ implies reconciling all of the samples holding CHC.
This strategy impairs performance compared to setting the threshold as we suggest in the following paragraph, on all of the datasets of up to 1.8\% (on OfficeHome) due to the existence of noise in the pseudo-labels, especially for small datasets. 

\begin{figure*}[t!]
\centering{\includegraphics[width=0.9\linewidth]{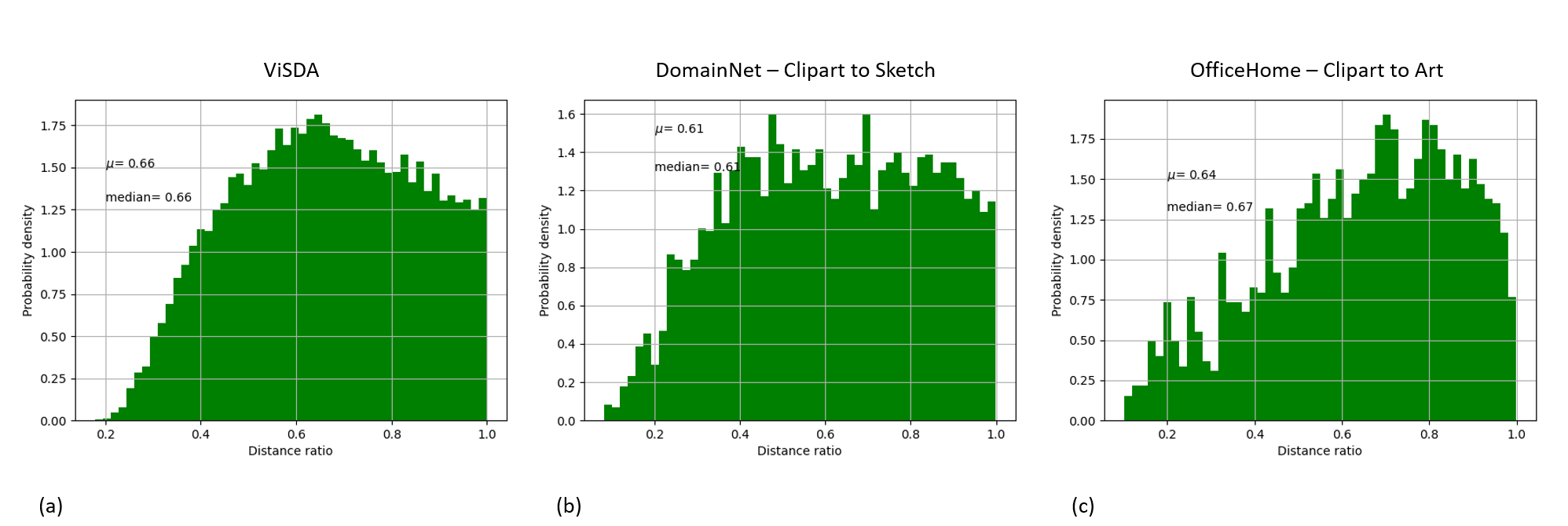}}
\caption{Uncertainty histograms of non-conflict target samples.}
\label{Fig:DistRatioHist}
\end{figure*}

For all of the domain shifts in our experiments, we find that the distributions of the uncertainty metric in Equation \ref{Eq:Ratio} of non-conflict target samples have similar mean and a median values around 0.65 at the beginning of the adaptation process.
Figure \ref{Fig:DistRatioHist} presents these distributions on three different domain shifts, one from each dataset. 
Therefore, we have an estimation for the median value of the distribution of the uncertainty metric 
for non-conflict samples. 
We choose the median as our uncertainty threshold since before adaptation both SR and PL are still noisy and the median is a safety margin.
In the following paragraph we show that our results are robust around this safety margin.

\begin{figure}[t!]
\centering{\includegraphics[width=0.9\columnwidth]{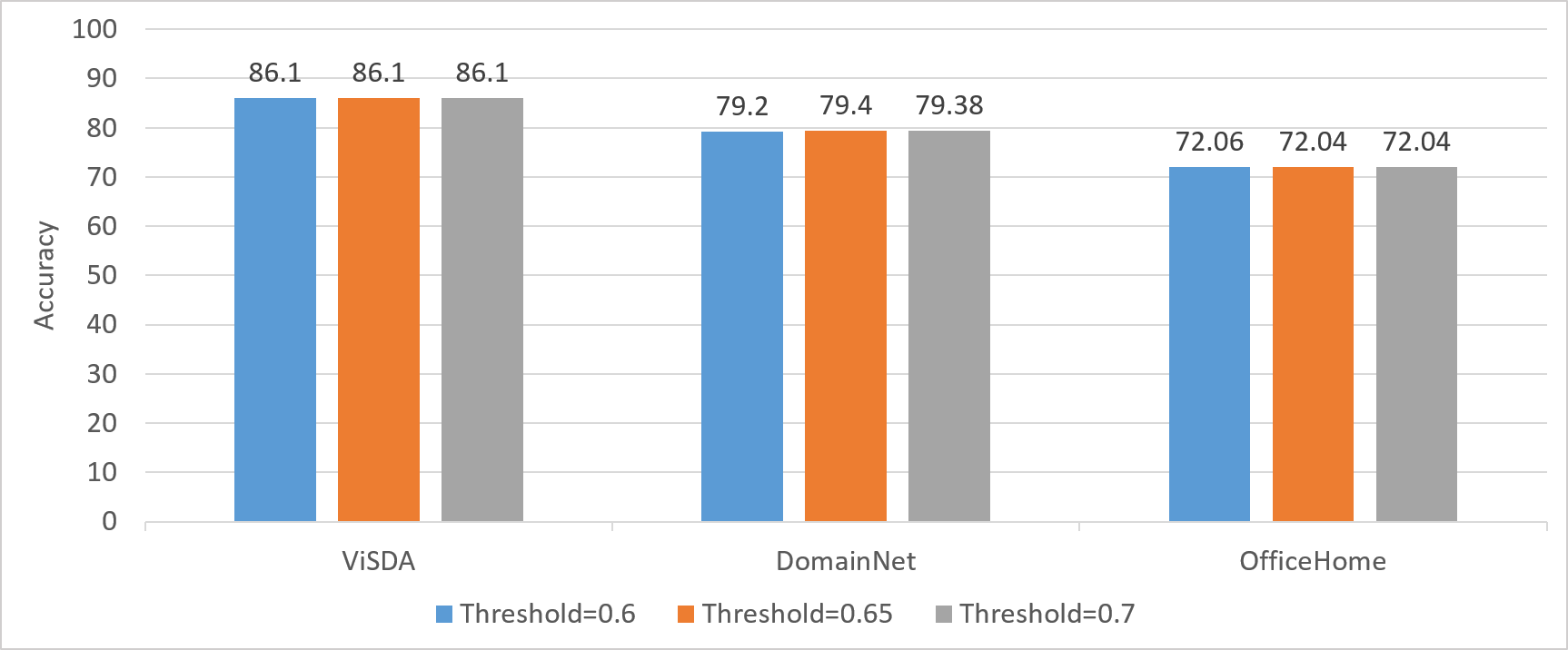}}
\caption{Threshold sensitivity performance comparison of RCHC for thresholds around 0.65 for all three datasets.}
\label{Fig:RatioThSensitivity}
\end{figure}

\paragraph{Uncertainty threshold sensitivity.}
Figure \ref{Fig:RatioThSensitivity} shows that the performance of RCHC is insensitive to threshold values around 0.65. Specifically, the figure presents performance comparisons of RCHC for three thresholds 0.6, 0.65 and 0.7 for all of the datasets.

\begin{figure*}[t!]
\centering{\includegraphics[width=0.9\textwidth]{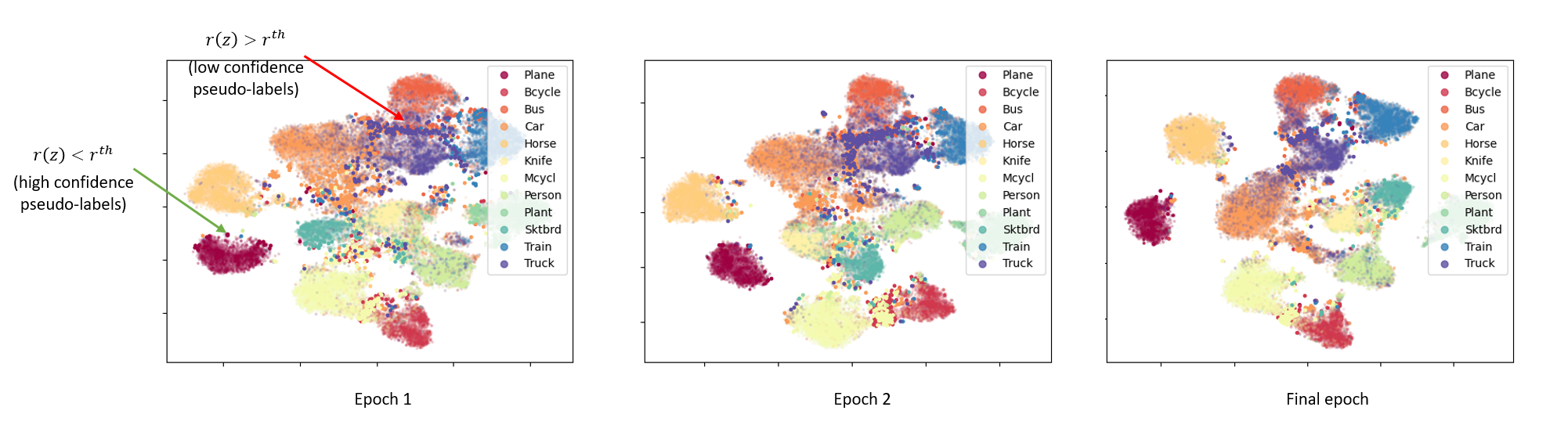}}
\caption{t-SNE for ViSDA at three different epochs. Opaque dots represent samples holding CHC and transparent dots represent non-conflict samples.}
\label{Fig:t-sne-visda}
\end{figure*}

\begin{figure*}[t!]
\centering{\includegraphics[width=0.9\textwidth]{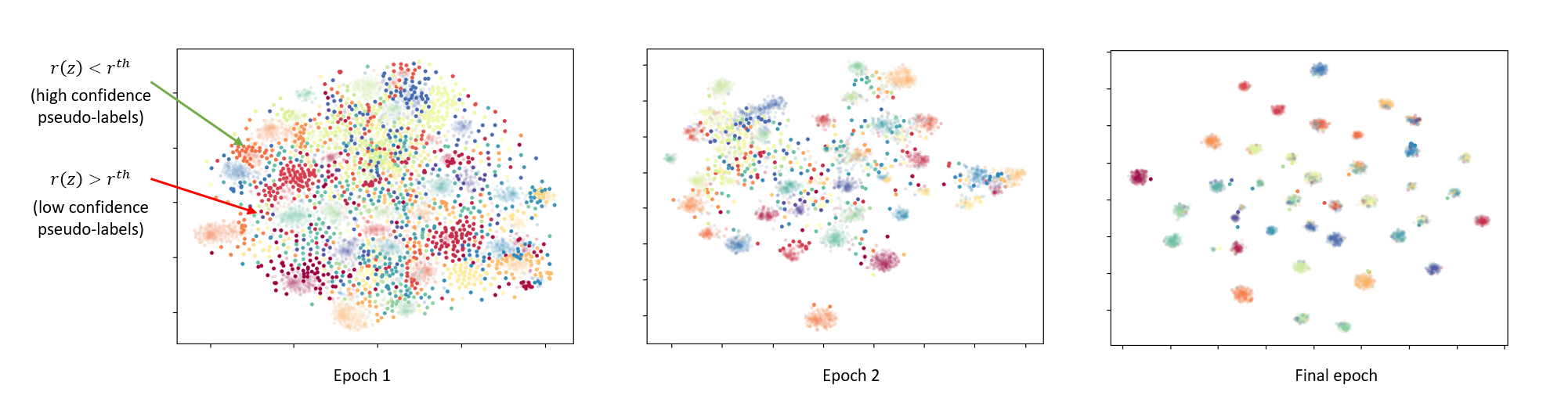}}
\caption{t-SNE for DomainNet Clipart to Sketch at three different epochs. Opaque dots represent samples holding CHC and transparent dots represent non-conflict samples.}
\label{Fig:t-sne-DomainNet}
\end{figure*}

\paragraph{t-SNE visualisations.}
Figures \ref{Fig:t-sne-visda} and \ref{Fig:t-sne-DomainNet} show t-SNE plots of target sample embeddings at three different epochs during target adaptation for ViSDA and DomainNet (Clipart to Sketch).
The figures confirm that samples holding CHC (marked with opaque dots) indeed mostly lay on the boundaries of class clusters and that they constitute a good set of candidate samples on which to reduce error accumulation during our SFDA strategy.
The intuition behind our uncertainty metric is that high certainty in the PL assignment is implied by closeness to a class centroid, however, samples have low certainty in the PL assignment if they lay close to the boundary of multiple classes simultaneously.

For example, in Figure \ref{Fig:t-sne-visda}, samples on the boundary between Truck and Bus have low certainty in the PL assignment, since they are on the boundary of multiple classes simultaneously, whereas samples on the boundary of Plane's cluster have high certainty in the PL assignments. Similar behavior is observed in Figure \ref{Fig:t-sne-DomainNet}.
The figures also show that the number of samples holding the centroid-hypothesis conflict is reduced during training and that at the end of the training the target samples are well clustered and separated.

\section{Discussion}

In this work, we introduce the centroid-hypothesis conflict (CHC), a conflict between self-supervised loss terms based on pseudo labels and entropy minimization used for SFDA.
We introduce RCHC, an SFDA method that reconciles CHC for samples where we have high certainty in the loss term implied by target pseudo-labels.

We demonstrate the effectiveness of RCHC on multiple datasets and its robustness across architectures including up-to-date backbones pre-trained on ImageNet-22K.
To the best of our knowledge, we are the first to present SFDA results with ConvNeXt for ViSDA, OfficeHome and DomainNet using variants of ConvNeXt (ConvNeXt-T and ConvNeXt-S), obtaining state-of-the-art results with model sizes comparable to those in previous SOTA results.

We believe that more research is required in order to improve the quality of pseudo-label assignments.
Since RCHC is agnostic to the strategy in which the centroids are generated, RCHC will benefit from improved clustering strategies and will integrate well with solutions such as the one proposed in \cite{sanqing2022BMD}.

{\small
\bibliographystyle{abbrv}
\bibliography{ref}
}

\end{document}